\documentclass[10pt,twocolumn,letterpaper]{article}
\usepackage{iccv}
\usepackage{times}
\usepackage{epsfig}
\usepackage{graphicx}
\usepackage{amsmath}
\usepackage{amssymb}
\usepackage{times}
\usepackage{epsfig}
\usepackage{graphicx}
\usepackage{amsmath}
\usepackage{amssymb}
\usepackage{authblk}
\usepackage{color}
\usepackage{times}
\usepackage{epsfig}
\usepackage{algorithm}
\usepackage{algorithmicx}
\usepackage{algpseudocode}
\usepackage{graphics}
\usepackage{threeparttable}
\usepackage{color}
\usepackage[normalem]{ulem}
\usepackage{multirow}
\usepackage{float}
\usepackage{amsfonts}
\usepackage{bm}
\usepackage{subfig}
\usepackage{array}
\usepackage[table]{xcolor}
\usepackage{colortbl}
\usepackage{pifont}
\usepackage{cite}
\usepackage{diagbox}
\usepackage{rotating}
\usepackage{booktabs}
\usepackage{overpic}
\usepackage{textcomp}
\usepackage{contour}
\usepackage{enumitem}
\usepackage{paralist}
\usepackage{makecell}
\usepackage{float}
\usepackage{cuted}
\usepackage{capt-of}
\usepackage{rotating}
\definecolor{mygray}{gray}{.92}

\makeatletter
\newcommand{\thickhline}{%
    \noalign {\ifnum 0=`}\fi \hrule height 1pt
    \futurelet \reserved@a \@xhline
}
\makeatother

\def\ourdataset{\textit{VACATION}} 

\usepackage[pagebackref=true,breaklinks=true,letterpaper=true,colorlinks,bookmarks=false]{hyperref}

\usepackage[pagebackref=true,breaklinks=true,letterpaper=true,colorlinks,bookmarks=false]{hyperref}

\iccvfinalcopy 
\def\iccvPaperID{421} 

\ificcvfinal\fi
\begin{document}

\title{Understanding Human Gaze Communication by \\ Spatio-Temporal Graph Reasoning}
\author{Lifeng Fan$^{1}\thanks{Lifeng Fan and Wenguan Wang contributed equally.}$~,~~\hspace{1pt}Wenguan Wang$^{2,1*}$,~~Siyuan Huang$^{1}$,~~Xinyu Tang$^{3}$,~~Song-Chun Zhu$^{1}$\hspace{1pt}   \\
	\small{$^1$} \small Center for Vision, Cognition, Learning and Autonomy, UCLA, USA \hspace{0pt}\\
	\small{$^2$} \small Inception Institute of Artificial Intelligence, UAE \hspace{0pt} \small{$^3$} \small University of Science and Technology of China, China \hspace{0pt}\\
   {\tt\small lfan@ucla.edu, {wenguanwang.ai@gmail.com}, {sczhu@stat.ucla.edu}}\\
   {\tt\small ~\url{https://github.com/LifengFan/Human-Gaze-Communication}}
	%
}

\maketitle
\begin{strip}\centering
\vspace{-65pt}
\includegraphics[width=\textwidth]{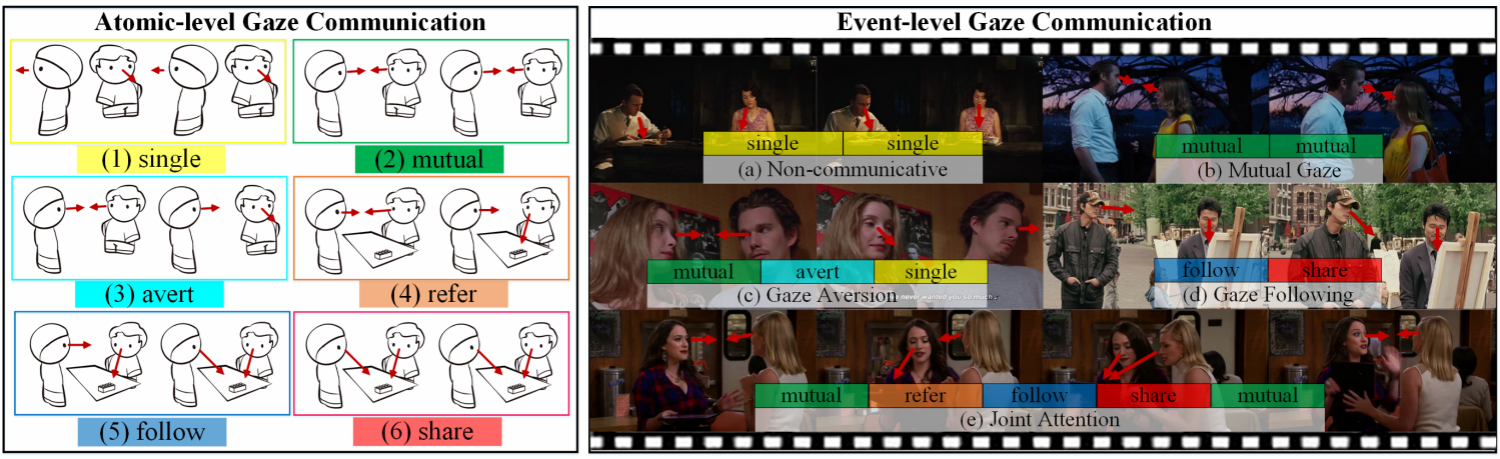}
\vspace{-20pt}
\captionof{figure}{\small We study human gaze communication dynamics in two hierarchical levels: atomic-level and event-level. Atomic-level gaze communication describes the fine-grained structures in human gaze interactions, \ie, \textit{single}, \textit{mutual}, \textit{avert}, \textit{refer}, \textit{follow} and \textit{share} (as shown in the left part). Event-level gaze communication refers to high-level, complex social communication events, including \textit{Non-communicative}, \textit{Mutual Gaze}, \textit{Gaze Aversion}, \textit{Gaze Following} and \textit{Joint Attention}. Each gaze communication event is a temporal composition of some atomic-level gaze communications (as shown in the right part).  
\label{fig:teaser}}
\vspace{-6pt}
\end{strip}

\thispagestyle{empty}

\begin{abstract}
This paper addresses a new problem of understanding human gaze communication in social videos from both atomic-level and event-level, which is significant for studying human social interactions. To tackle this novel and challenging problem, we contribute a large-scale video dataset, \ourdataset, which covers diverse daily social scenes and gaze communication behaviors with complete annotations of objects and human faces, human attention, and communication structures and labels in both atomic-level and event-level. Together with \ourdataset, we propose a spatio-temporal graph neural network to explicitly represent the diverse gaze interactions in the social scenes and to infer atomic-level gaze communication by message passing. We further propose an event network with encoder-decoder structure to predict the event-level gaze communication. Our experiments demonstrate that the proposed model improves various baselines significantly in predicting the atomic-level and event-level gaze communications.    

\end{abstract}

\vspace{-16pt}
\section{Introduction}
\vspace{-4pt}

In this work, we introduce the task of understanding human \textit{gaze communication} in social interactions. Evidence from psychology suggests that eyes are a cognitively special stimulus, with unique ``hard-wired'' pathways in the brain dedicated to their interpretation and humans have the unique ability to infer others' intentions from eye gazes~\cite{emery2000eyes}. Gaze communication is a primitive form of human communication, whose underlying social-cognitive and social-motivational infrastructure acted as a psychological platform on which various linguistic systems could be built~\cite{tomasello2010origins}. Though verbal communication has become the primary form in social interaction, gaze communication still plays an important role in conveying hidden mental state and augmenting verbal communication~\cite{admoni2017socialeyegaze}. To better understand human communication, we not only need natural language processing (NLP), but also require a systematical study of human gaze communication mechanism.

The study of human gaze communication in social interaction is essential for the following several reasons: 1) it helps to better understand multi-agent gaze communication behaviors in realistic social scenes, especially from social and psychological views; 2) it provides evidences for robot systems to learn human behavior patterns in gaze communication and further facilitates intuitive and efficient interactions between human and robot; 3) it enables simulation of more natural human gaze communication behaviors in Virtual Reality environment; 4) it builds up a common sense knowledge base of human gaze communication for studying human mental state in social interaction; 5) it helps to evaluate and diagnose children with autism.

Over the past decades, lots of research~\cite{haith1977eye,kobayashi1997unique,itier2009neural,jording2018social} on the types and effects of social gazes have been done in cognitive psychology and neuroscience communities. With previous efforts and established terminologies, we distinguish atomic-level gaze communications into six classes:

\noindent{\small\textbullet}~\textit{Single} refers to individual gaze behavior without any social communication intention (see Fig.~\ref{fig:teaser} (1)).

\noindent{\small\textbullet}~\textit{Mutual}~\cite{admoni2017socialeyegaze, argyle1976gaze} gaze occurs when two agents look into eyes of each other (see Fig.~\ref{fig:teaser} (2)), which is the strongest mode of establishing a communicative link between human agents. \textit{Mutual} gaze can capture attention, initialize a conversation, maintain engagement, express feelings of trust and extroversion, and signal availability for interaction in cases like passing objects to a partner. 

\noindent{\small\textbullet}~\textit{Avert}~\cite{riemer1949avert, glenberg1998averting} refers to averted gaze and happens when gaze of one agent is shifted away from another in order to avoid mutual gaze (see Fig.~\ref{fig:teaser} (3)). \textit{Avert} gaze expresses distrust, introversion, fear, and can also modulate intimacy, communicate thoughtfulness or signal cognitive effort such as looking away before responding to a question. 

\noindent{\small\textbullet}~\textit{Refer}~\cite{senju2006development} means referential gaze and happens when one agent tries to induce another agent's attention to a target via gaze (see Fig.~\ref{fig:teaser} (4)). Referential gaze shows intents to inform, share or request sth. We can use \textit{refer} gaze to eliminate uncertainty about reference and respond quickly. 

\noindent{\small\textbullet}~\textit{Follow}~\cite{shepherd2010gazefollowing, zuberbuhler2008gaze, brooks2005development} means following gaze and happens when one agent perceives gaze from another and follows to contact with the stimuli the other is attending to (see Fig.~\ref{fig:teaser} (5)). Gaze following is to figure out partner's intention.

\noindent{\small\textbullet}~\textit{Share}~\cite{okamoto2006development} means shared gaze and appears when two agents are gazing at the same stimuli (see Fig.~\ref{fig:teaser} (6)).  

The above atomic-level gazes capture the most general, core and fine-grained gaze communication patterns in human social interactions. We further study the long-term, coarse-grained temporal compositions of the above six atomic-level gaze communication patterns, and generalize them into totally five gaze communication events, \ie, \textit{Non-communicative}, \textit{Mutual Gaze}, \textit{Gaze Aversion}, \textit{Gaze Following} and \textit{Joint Attention}, as illustrated in the right part of Fig~\ref{fig:teaser}. Typically the temporal order of atomic gazes means different phases of each event. \textit{Non-communicative} (see Fig.~\ref{fig:teaser} (a)) and \textit{Mutual Gaze} (see Fig.~\ref{fig:teaser} (b)) are one-phase events and simply consist of \textit{single} and \textit{mutual} respectively. \textit{Gaze Aversion} (see Fig.~\ref{fig:teaser} (c)) starts from \textit{mutual}, then \textit{avert} to \textit{single}, demonstrating the avoidance of mutual eye contact. \textit{Gaze Following} (see Fig.~\ref{fig:teaser} (d)) is composed of \textit{follow} and \textit{share}, but without \textit{mutual}, meaning that there is only one-way awareness and observation, no shared attention nor knowledge. \textit{Joint Attention} (see Fig.~\ref{fig:teaser} (e)) is the most advanced and appears when two agents have the same intention to share attention on a common stimuli and both know that they are sharing something as common ground. Such event consists of several phases, typically beginning with \textit{mutual} gaze to establish communication channel, proceeding to \textit{refer} gaze to draw attention to the target, and \textit{follow} gaze to check the referred stimuli, and cycling back to \textit{mutual} gaze to ensure that the experience is shared~\cite{moore2014joint}. Clearly, recognizing and understanding atomic-level gaze communication patterns is necessary and significant first-step for comprehensively understanding human gaze behaviors.

To facilitate the research of gaze communication understanding in computer vision community, we propose a large-scale social video dataset named \ourdataset~(Video gAze CommunicATION) with complete gaze communication annotations. With our dataset, we aim to build spatio-temporal attention graph given a third-person social video sequence with human face and object bboxes, and predict gaze communication relations for this video in both atomic-level and event-level. Clearly, this is a structured task that requires a comprehensive modeling of human-human and human-scene interactions in both spatial and temporal domains.

Inspired by recent advance in graph neural network~\cite{Qi_2018_ECCV, velickovic2018graph}, we propose a novel spatio-temporal reasoning graph network for atomic-level gaze communication detection as well as an event network with encoder-decoder structure for event-level gaze communication understanding. The reasoning model learns the relations among social entities and iteratively propagates information over a social graph. The event network utilizes the encoder-decoder structure to eliminate the noises in gaze communications and learns the temporal coherence for each event to classify event-level gaze communication. 

This paper makes \textbf{three major contributions}:
\begin{compactitem}
\item [1)] It proposes and addresses a new task of gaze communication learning in social interaction videos. To the best of our knowledge, this is the first work to tackle such problem in computer vision community.
\item [2)] It presents a large-scale video dataset, named \ourdataset, covering diverse social scenes with complete gaze communication annotations and benchmark results for advancing gaze communication study.
\item [3)] It proposes a spatio-temporal graph neural network and an event network to hierarchically reason both atomic- and event-level gaze communications in videos.
\end{compactitem}

\vspace{-4pt}
\section{Related Work}

\vspace{-4pt}
\subsection{Gaze Communication in HHI} \label{sec:NCHHI}
\vspace{-4pt}

Eye gaze is closely tied to underlying attention, intention, emotion and personality~\cite{kleinke1986gaze}. Gaze communication allows people to communicate with one another at the most basic level regardless of their familiarity with the prevailing verbal language system. Such social eye gaze functions thus transcend cultural differences, forming a universal language~\cite{burgoon2016nonverbal}. During conversations, eye gaze can be used to convey information, regulate social intimacy, manage turn-taking, control conversational pace, and convey social or emotional states~\cite{kleinke1986gaze}. People are also good at identifying the target of their partner's referential gaze and use this information to predict what their partner is going to say~\cite{staudte2011investigating,boucher2012reach}.


In a nutshell, gaze communication is omnipresent and multifunctional~\cite{burgoon2016nonverbal}. Exploring the role of gaze communication in HHI is an essential research subject, but it has been rarely touched by computer vision researchers. Current research in computer vision community~\cite{itti1998model,borji2013state,wang2018deep,fan2019shifting,wang2019salient} mainly focuses on studying the salient properties of the natural environment to model human visual attention mechanism. Only a few~\cite{park20123d, soo2015social, fan2018inferring} studied human shared attention behaviors in social scenes.

\vspace{-4pt}
\subsection{Gaze Communication in HRI} \label{sec:NCHRI}
\vspace{-4pt}

To improve human-robot collaboration, the field of HRI strives to develop effective gaze communication for robots~\cite{admoni2017socialeyegaze}. Researchers in robotics tried to incorporate responsive, meaningful and convincing eye gaze into HRI~\cite{admoni2014data, andrist2015look}, which helps the humanoid agent to engender the desired familiarity and trust, and makes HRI more intuitive and fluent. Their efforts vary widely~\cite{srinivasan2011survey,andrist2014conversational,admoni2017socialeyegaze}, including human-robot visual dialogue interaction~\cite{mutlu2012conversational,staudte2009visual,liu2012generation}, storytelling~\cite{mutlu2006storytelling}, and socially assistive robotics~\cite{tapus2007socially}. For example, a tutoring or assistive robot can demonstrate attention to and engagement with the user by performing proper \textit{mutual} and \textit{follow} gazes~\cite{meltzoff2010social}, direct user attention to a target using \textit{refer} gaze, and form joint attention with humans~\cite{huang2011effects}. A collaborative assembly-line robot can also enable object reference and joint attention by gazes. Robots can also serve as therapy tools for children with autism.

\begin{figure*}[t]
	\centering
   \includegraphics[width=1\linewidth]{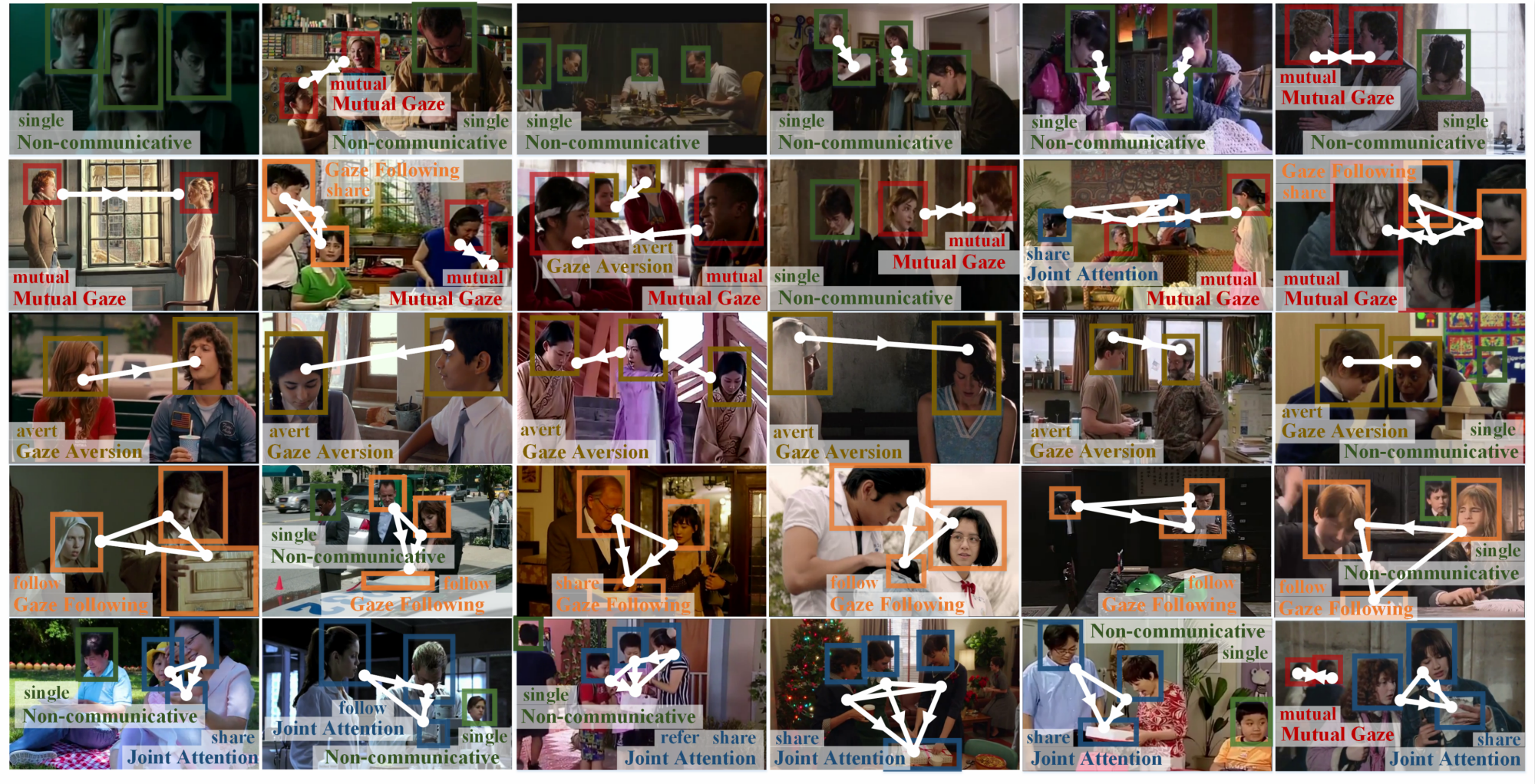}
   \vspace*{-18pt}
   \caption{\small \textbf{Example frames and annotations of our \ourdataset~dataset}, showing that our dataset covers rich gaze communication behaviors, diverse general social scenes, different cultures, \etc. It also provides rich annotations, \ie, human face and object bboxes, gaze communication structures and labels. Human faces and related objects are marked by boxes with the same color of corresponding communication labels. White lines link entities with gaze relations in a temporal sequence and white arrows indicate gaze directions in the current frame. There may exist various number of agents, many different gaze communication types and complex communication relations in one frame, resulting in a highly-challenging and structured task. See \S\ref{sec:dataset} for details.}
\label{fig:example}
\vspace*{-10pt}
\end{figure*}

\vspace{-3pt}
\subsection{Graph Neural Networks} \label{sec:GNN}
\vspace{-3pt}
Recently, graph neural networks~\cite{scarselli2009graph,li2016gated,jain2016structural,gilmer2017neural} received increased interests since they inherit the complementary advantages of graphs (with strong representation ability) and neural networks (with end-to-end learning power). These models typically pass local messages on graphs to explicitly capture the relations among nodes, which are shown to be effective at a large range of structured tasks, such as graph-level classification~\cite{bruna2014spectral,dai2016discriminative,velickovic2018graph}, node-level classification~\cite{hamilton2017inductive}, relational reasoning~\cite{santoro2017simple,kipf2018neural}, multi-agent communications~\cite{sukhbaatar2016learning,battaglia2016interaction}, human-object interactions~\cite{Qi_2018_ECCV,fang2018pairwise}, and scene understanding~\cite{marino2017more, li2017situation}. Some others~\cite{duvenaud2015convolutional,niepert2016learning,kipf2017semi,simonovsky2017dynamic,Chen_2018_CVPR} tried to generalize convolutional architecture over graph-structured data. Inspired by above efforts, we build a spatio-temporal social graph to explicitly model the rich interactions in dynamic scenes. Then a spatio-temporal reasoning network is proposed to learn gaze communications by passing messages over the social graph.

\vspace{-3pt}
\section{The Proposed \ourdataset~Dataset}\label{sec:dataset}
\vspace{-3pt}

\ourdataset~contains 300 social videos with diverse gaze communication behaviors. Example frames can be found in Fig.~\ref{fig:example}. Next we will elaborate \ourdataset~from the following essential aspects.

\vspace{-3pt}
\subsection{Data Collection}
\vspace{-3pt}
Quality and diversity are two essential factors considered in our data collection.

\begin{table}[t!]
  \centering
  \resizebox{0.49\textwidth}{!}{
    \setlength\tabcolsep{2pt}
    \renewcommand\arraystretch{1.0}
  \begin{tabular}{c|c||c|c|c|c|c}
  \hline\thickhline
     \rowcolor{mygray}
   \multicolumn{2}{c||}{\textbf{Event-}}&\makecell{\textit{Non-Comm.}} &\makecell{\textit{Mutual Gaze}} & \makecell{\textit{Gaze Aversion}} & \makecell{\textit{Gaze Following}} &\makecell{\textit{Joint Attention}} \\
   \rowcolor{mygray}
   \multicolumn{2}{c||}{\textbf{level (\%)}}&28.16  &24.00 &10.00 &10.64 & 27.20\\
  \hline
   \hline 
   \multirow{6}{*}{\rotatebox{90}{\textbf{Atomic-level(\%)}}}  &\makecell{\textit{single}}  &92.20 &15.99 &3.29 &39.26 &26.91\\
   &\makecell{\textit{mutual}}  &0.76 &75.64 &14.15 &0.00 & 16.90\\
   &\makecell{\textit{avert}}  &1.34  &6.21 &81.71 &0.00 &1.18\\
   &\makecell{\textit{refer}}  &0.00  &0.37 &0.15 & 0.62 &7.08\\
   &\makecell{\textit{follow}}  &1.04  &0.29 &0.00 & 10.71 & 2.69\\
   &\makecell{\textit{share}}  &4.66  &1.50 &0.70 &49.41 &45.24\\
  \hline
  \end{tabular}
  }
  \vspace*{-6pt}
  \caption{\small\textbf{Statistics of gaze communication categories} in our \ourdataset~dataset, including the distribution of event-level gaze communication category over full dataset and the distribution of atomic-level gaze communication for each event-level category. 
  }\label{tab:DEG}
  \vspace*{-15pt}
\end{table}

\noindent\textbf{High quality}. We searched the Youtube engine for more than 50 famous TV shows and movies (\eg, The Big Bang Theory, Harry Potter, \etc). Compared with self-shot social data in laboratory or other limited environments, these stimuli provide much more natural and richer social interactions in general and representative scenes, and are closer to real human social behaviors, which helps to better understand and model real human gaze communication behaviors. After that, about $1, 000$  video clips are roughly split from the retrieved results. We further eliminate the videos with big logo or of low-quality. Each of the rest videos is then cropped with accurate shot boundaries and uniformly stored in MPEG-4 format with $640\!\times\! 360$ spatial resolution. \ourdataset~finally comprises a total of 300 high-quality social video sequences with 96,993 frames and 3,880-second duration. The lengths of videos span from 2.2 to 74.56 seconds and are 13.28 seconds on average.

\noindent\textbf{Diverse social scenes}. The collected videos cover diverse daily social scenes (\eg, party, home, office, \etc), with different cultures (\eg, American, Chinese, Indian, \etc). The appearances of actors/actresses, costume and props, and scenario settings, also vary a lot, which makes our dataset more diverse and general. By training on such data, algorithms are supposed to have better generalization ability in handling diverse realistic social scenes. 

\vspace{-3pt}
\subsection{Data Annotation and Statistics}\label{sec:DA}
\vspace{-3pt}
Our dataset provides rich annotations, including human face and object bounding boxes, human attention, atomic-level and event-level gaze communication labels. The annotation takes about 1,616 hours in total, considering an average annotation time of 1 minute per frame. Three extra volunteers are included in this process. 

\noindent\textbf{Human face and object annotation}. We first annotate each frame with bounding boxes of human face and key object, using the online video annotation platform Vatic~\cite{vondrick2013efficiently}. 206,774 human face bounding boxes (avg. 2.13 per frame) and 85,441 key object bounding boxes (avg. 0.88 per frame) are annotated in total. 

\noindent\textbf{Human attention annotation}. We annotate the attention of each person in each frame, \ie the bounding box (human face or object) this person is gazing at.  

\noindent\textbf{Gaze communication labeling}. The annotators are instructed to annotate both atomic-level and event-level gaze communication labels for every group of people in each frame. To ensure the annotation accuracy, we used cross-validation in the annotation process, \ie, two volunteers annotated all the persons in the videos separately, and the differences between their annotations were judged by a specialist in this area. See Table~\ref{tab:DEG} for the information regarding the distributions of gaze communication categories.

\begin{table}[t!]
  \centering
  \resizebox{0.45\textwidth}{!}{
    \setlength\tabcolsep{8pt}
    \renewcommand\arraystretch{1.0}
  \begin{tabular}{r||c|c|c|c}
  \hline\thickhline
  \rowcolor{mygray}
   \multirow{-1}{*}{\small\textbf{\ourdataset}}&\multirow{-1}{*}{\# Video}&\multirow{-1}{*}{\# Frame}&\multirow{-1}{*}{\# Human} &\multirow{-1}{*}{\# GCR}\\
  \hline
  \hline
  \textit{training}  &180  &57,749  &123,812  &97,265\\
  \textit{validation}  &60  &22,005  &49,012 &42,066 \\
  \textit{testing}  &60  &17,239  &33,950 &25,034  \\
  \hline
  \textit{full dataset}  &300  &96,993  &206,774 &164,365 \\
  \hline
  \end{tabular}
  }
  \vspace*{-6pt}
  \caption{\small\textbf{Statistics of dataset splitting}. GCR refers to Gaze Communication Relation. See \S\ref{sec:DA} for more details.}\label{tab:DS}
  \vspace*{-20pt}
\end{table}


\noindent\textbf{Dataset splitting}. Our dataset is split into training, validation and
testing sets with the ratio of 6:2:2. We arrive at a unique split consisting of 180 training (57,749
frames), 60 validation (22,005 frames), and 60 testing videos (17,239
frames). To avoid over-fitting, there is no source-overlap among videos in different sets (see Table~\ref{tab:DS} for more details). 

\vspace{-3pt}
\section{Our Approach}
\vspace{-3pt}

We design a spatio-temporal graph neural network to explicitly represent the diverse interactions in social scenes and infer atomic-level gaze communications by passing messages over the graph. Given the atomic-level gaze interaction inferences, we further design an event network with encoder-decoder structure for event-level gaze communication reasoning.
As shown in Fig.~\ref{fig:overview}, gaze communication entities, \ie, human, social scene, are represented by graph nodes, gaze communication structures are represented by edges. We introduce notations and formulations in \S\ref{sec:formulation} and provide more implementation details in \S\ref{sec:id}.

\vspace{-3pt}
\subsection{Model Formulation}\label{sec:formulation}
\vspace{-3pt}
\noindent\textbf{Social Graph.} We first define a social graph as a \textit{complete graph} $\mathcal{G}\!=\!(\mathcal{V},\mathcal{E})$, where node $v\!\in\!\mathcal{V}$ takes unique value from $\{1,\cdots,|\mathcal{V}|\}$, representing the entities (\ie, scene, human) in social scenes, and edge $e\!=\!(v,w)\in\mathcal{E}$ indicates a directed edge $v\!\rightarrow\!w$, representing all the possible human-human gaze interactions or human-scene relations. There is a special node $s\!\in\!\mathcal{V}$ representing the social scene. For node $v$, its \textit{node representation}/\textit{embedding} is denoted by a $V$-dimensional vector: $\mathbf{x}_v\!\in\!\mathbb{R}^{V\!}$. Similarly, the \textit{edge representation/embedding} for edge $e\!=\!(v,w)$ is denoted by an $E$-dimensional vector: $\mathbf{x}_{v,w\!}\!\in\!\mathbb{R}^E$. Each human node $v\!\in\!\mathcal{V}\backslash s$ has an output state $l_v\!\in\!\mathcal{L}$ that takes a value from a set of atomic gaze labels: $\mathcal{L}\!=\!\{$\textit{single}, \textit{mutual}, \textit{avert}, \textit{refer}, \textit{follow}, \textit{share}$\}$. We further define an
adjacency matrix $\textbf{A}\!\in\![0,1]^{|\mathcal{V}|\times |\mathcal{V}|}$ to represent the communication structure over our complete social graph $\mathcal{G}$, where each element $a_{v,w}$ represents the connectivity from node $v$ to $w$.

Different from most previous graph neural networks that only focus on inferring graph- or node-level labels, our model aims to learn the graph structure $\textbf{A}$ and the visual labels $\{l_v\}_{v\in\mathcal{V}\backslash s}$ of all the human nodes $\mathcal{V}\backslash s$ simultaneously.

\begin{figure}[t]
\centering
   \includegraphics[width=0.99\linewidth]{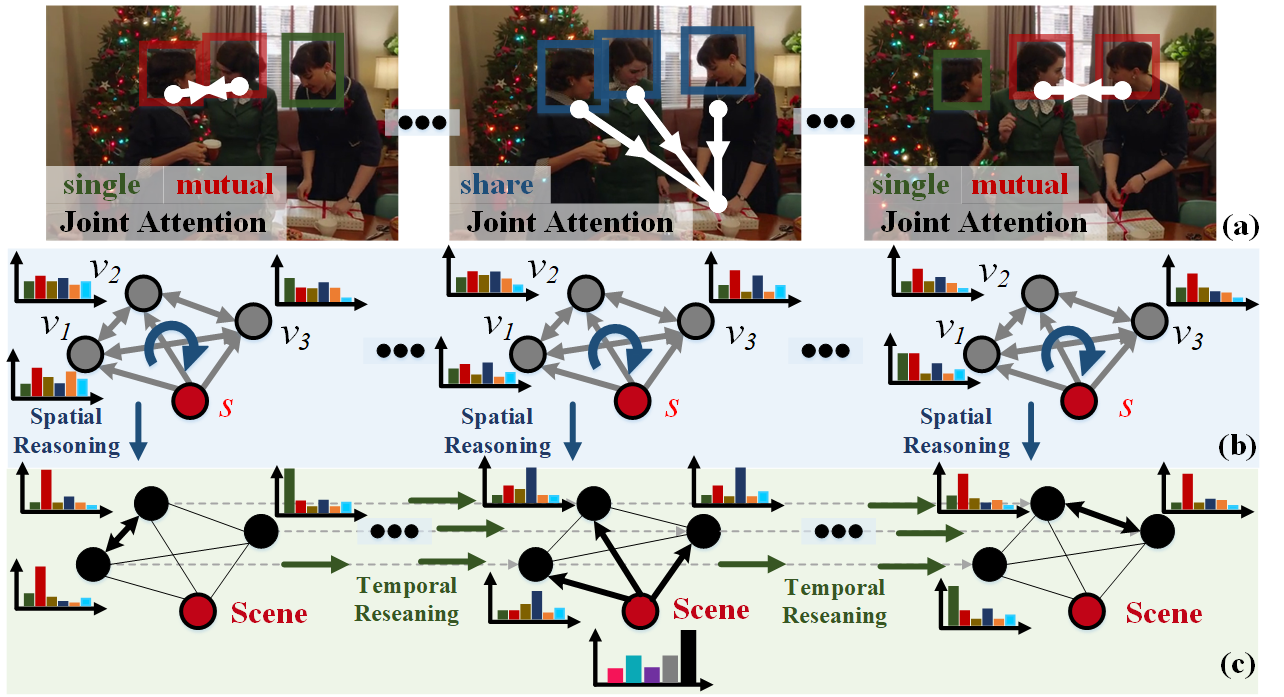}
   \vspace*{-6pt}
   \caption{\small\textbf{Illustration of the proposed spatio-temporal reasoning model} for gaze communication understanding. Given an input social video sequence (a), for each frame, a spatial reasoning process (b) is first performed for simultaneously capturing gaze communication relations (social graph structure) and updating node representations through message propagation. Then, in (c), a temporal reasoning process is applied for each node to dynamically update node representation over temporal domain, which is achieved by an LSTM. Bolder edges represent higher connectivity weight inferred in spatial reasoning step (b). See \S\ref{sec:formulation} for details.}
   \vspace*{-10pt}
\label{fig:overview}
\end{figure}

\begin{figure*}[t]
\centering
   \includegraphics[width=0.99\linewidth]{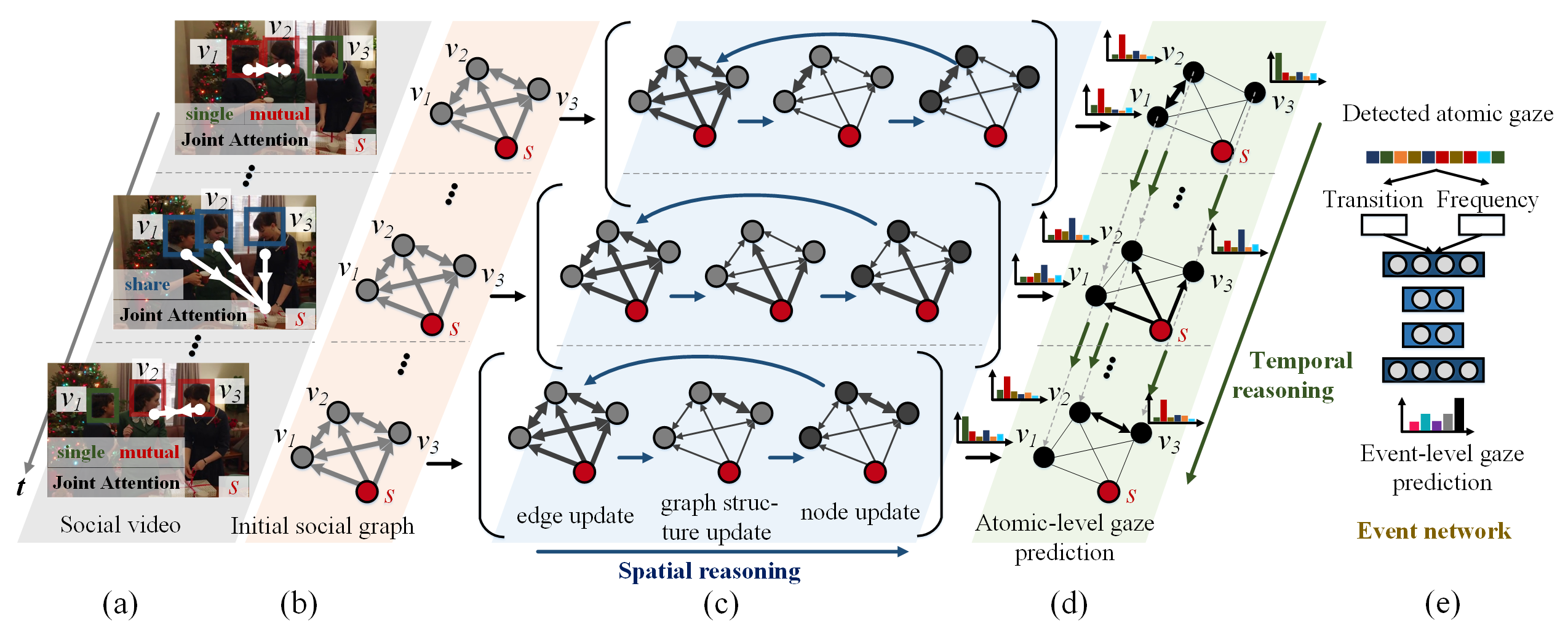}
   \vspace*{-10pt}
   \caption{\small\textbf{Detailed architecture of the proposed spatio-temporal reasoning model} for gaze communication understanding. See the last paragraph in \S\ref{sec:formulation} for detailed descriptions.}
   \vspace*{-13pt}
\label{fig:model}
\end{figure*}

To this end, our spatio-temporal reasoning model is designed to have two steps. First, in spatial domain, there is a message passing step (Fig.~\ref{fig:overview} (b)) that iteratively learns gaze communication structures $\textbf{A}$ and propagates information over $\textbf{A}$ to update node representations. Second, as shown in Fig.~\ref{fig:overview} (c), an LSTM is incorporated into our model for more robust node representation learning by considering temporal dynamics. A more detailed model architecture is schematically depicted in Fig.~\ref{fig:model}. In the following, we describe the above two steps in detail.

\noindent\textbf{Message Passing based Spatial Reasoning.}  Inspired by previous graph neural networks~\cite{gilmer2017neural,Qi_2018_ECCV,kipf2018neural}, our message passing step is designed to have three phases, an \textit{edge update} phase, a \textit{graph structure update} phase, and a \textit{node update} phase.
The whole message passing process runs for $N$ iterations to iteratively propagate information. In $n$-th iteration step, we first perform the edge update phase that updates edge representations $\mathbf{y}^{(n)}_{v,w}$ by collecting information from connected nodes:
\begin{equation}\small
    \begin{aligned}
    \mathbf{y}^{(n)}_{v,w}= f_E(\langle\mathbf{y}^{(n\!-\!1)}_{v},\mathbf{y}^{(n\!-\!1)}_{w},\mathbf{x}_{v,w}\rangle),
    \end{aligned}\label{eq:1}
\end{equation}
where $\mathbf{y}^{(n\!-\!1)}_{v}$ indicates the node representation of $v$ in ${(n\!-\!1)}$-th step, and $\langle\cdot,\cdot\rangle$ denotes concatenation of vectors. $f_E$ represents an \textit{edge update function} $f_E\!:\! \mathbb{R}^{2V+E}\!\rightarrow\! \mathbb{R}^{E}$, which is implemented by a neural network.

After that, the graph structure update phase updates the adjacency matrix $\textbf{A}$ to infer the current social graph structure, according to the updated edge representations $\mathbf{y}^{(n)}_{v,w}$:
\begin{equation}\small
    \begin{aligned}
    a^{(n)}_{v,w}= \sigma(f_A(\mathbf{y}^{(n)}_{v,w})),
    \end{aligned}\label{eq:2}
\end{equation}
where the connectivity matrix $\textbf{A}^{(n)} = [a^{(n)}_{v,w}]_{v,w}$ encodes current visual communication structures. $f_A\!:\! \mathbb{R}^{E}\!\rightarrow\! \mathbb{R}$ is a \textit{connectivity readout network} that maps an edge representation into the connectivity weight, and $\sigma$ denotes nonlinear activation function.

Finally, in the node update phase, we update node representations $\mathbf{y}^{(n)}_{v}$ via considering all the incoming edge information weighted by the corresponding connectivity:
\begin{equation}\small
    \begin{aligned}
    \mathbf{y}^{(n)}_{v}=f_V(\langle\sum\nolimits_{w}\!\!a^{(n)}_{v,w}\mathbf{y}^{(n)}_{v,w},\mathbf{x}_{v}\rangle),
    \end{aligned}\label{eq:3}
\end{equation}
where $f_{V\!\!}:\! \mathbb{R}^{\!V\!+\!E}\!\!\rightarrow\!\! \mathbb{R}^{\!V\!\!}$ represents a \textit{node update network} .

The above functions $f(\cdot)$ are all learned differentiable functions. In the above message passing process,  we infer social communication structures in the graph structure update phase (Eq.~\ref{eq:2}), where the relations between each social entities are learned through updated edge representations (Eq.~\ref{eq:1}). Then, the information is propagated through the learned social graph structure and the hidden state of each node is updated based on its history and incoming messages from its neighborhoods (Eq.~\ref{eq:3}). If we know whether there exist interactions between nodes (human, object), \ie, given the groundtruth of $\textbf{A}$, we can learn $\textbf{A}$ in an \textit{explicit} manner, which is similar to the graph parsing network~\cite{Qi_2018_ECCV}. Otherwise, the adjacent matrix $\textbf{A}$ can be viewed as an attention or gating mechanism that automatically weights the messages and can be learned in an \textit{implicit} manner; this shares a similar spirit with graph attention network~\cite{velickovic2018graph}. More implementation details can be found in \S\ref{sec:id}.

 \noindent\textbf{Recurrent Network based Temporal Reasoning.} Since our task is defined on a spatio-temporal domain, temporal dynamics should be considered for more comprehensive reasoning. With the updated human node representations $\{\mathbf{y}_v\!\in\!\mathbb{R}^V\}_{v\in\mathcal{V}\backslash s}$ from our message passing based spatial reasoning model, we further apply LSTM to each node for temporal reasoning. More specifically, our temporal reasoning step has two phases:  a \textit{temporal message passing} phase and a \textit{readout} phase. We denote by $\mathbf{y}^t_v$ the feature of a human node $v\in\mathcal{V}\backslash s$ at time $t$, which is obtained after $N$-iteration spatial message passing. In the temporal message passing phase, we propagate the information over the temporal axis using LSTM:
 \begin{equation}\small
     \begin{aligned}
     \mathbf{h}^t_{v}= f_{\text{LSTM}}(\mathbf{y}^t_{v} | \mathbf{h}^{t\!-\!1}_{v}),
     \end{aligned}\label{eq:4}
 \end{equation}
 where $f_{\text{LSTM}}\!:\! \mathbb{R}^{V}\!\rightarrow\! \mathbb{R}^{V}$ is an LSTM based temporal reasoning function that updates the node representation using temporal information. $\mathbf{y}^t_{v}$ is used as the input of the LSTM at time $t$, and $\mathbf{h}^t_{v}$ indicates the corresponding hidden state output via considering previous information $\mathbf{h}^{t-1}_{v}$. 

 Then, in the readout phase, for each human node $v$, a corresponding gaze label $\hat{l}^t_v\!\in\!\mathcal{L}$ is predicted from the final node representation $\mathbf{h}^t_{v}$:
 \begin{equation}\small
     \begin{aligned}
     \hat{l}^t_v= f_{R}(\mathbf{h}^t_{v}),
     \end{aligned}\label{eq:5}
 \end{equation}
 where $f_{R}\!:\! \mathbb{R}^{V}\!\rightarrow\!\mathcal{L}$ maps the node feature into the label space
 $\mathcal{L}$, which is implemented by a classifier network.
 
\noindent\textbf{Event Network.} The event network is designed with an encoder-decoder structure to learn the correlation of the atomic gazes and classify the event-level gaze communication for each video sequence. To reduce the large variance of video length, we pre-process the input atomic gaze sequence into two vectors: i) the transition vector that records each transition from one category of atomic gaze to another, and ii) the frequency vector that computes the frequency of each atomic type. The encoder individually encodes the transition vector and frequency vector into two embedded vectors. The decoder decodes the concatenation of these two embedded vectors and makes final event label prediction. Since the atomic gaze communications are noisy within communicative activities, the encoder-decoder structure will try to eliminate the noise and improve the prediction performance. The encoder and decoder are both implemented by fully-connected layers.

\begin{table*}[t]
  \centering
  \resizebox{0.99\textwidth}{!}{
    \setlength\tabcolsep{1pt}
    \renewcommand\arraystretch{1}
  \begin{tabular}{r||c|c|c|c|c|c|c|c|c|c|c|c|c|c}
  \hline\thickhline
  \rowcolor{mygray}
     & \multicolumn{14}{c}{Atomic-level Gaze Communication (Precision  \&   F1-score)}\\
     \cline{2-15}
     \rowcolor{mygray}
   \multirow{-2}{*}{Task~~~~~}&\multicolumn{2}{c}{\textit{single}}&\multicolumn{2}{c}{\textit{mutual}}
   &\multicolumn{2}{c}{\textit{avert}}&\multicolumn{2}{c}{\textit{refer}}&\multicolumn{2}{c}{\textit{follow}}
   &\multicolumn{2}{c}{\textit{share}}&\multicolumn{2}{c}{\textit{Avg. Acc.}}\\
   \hline
  Metric~~~~ &$\mathcal{P}$ (\%) $\uparrow$ &$\mathcal{F}$ (\%) $\uparrow$  &$\mathcal{P}$ (\%)  $\uparrow$ &$\mathcal{F}$ (\%) $\uparrow$ &$\mathcal{P}$ (\%) $\uparrow$ &$\mathcal{F}$ (\%) $\uparrow$ &$\mathcal{P}$ (\%) $\uparrow$ &$\mathcal{F}$ (\%) $\uparrow$ &$\mathcal{P}$ (\%) $\uparrow$ &$\mathcal{F}$ (\%) $\uparrow$ &$\mathcal{P}$ (\%) $\uparrow$ &$\mathcal{F}$ (\%) $\uparrow$ & top-1 (\%) $\uparrow$ & top-2 (\%) $\uparrow$\\
  \hline
  \hline
  \textbf{Ours-full}   &\multirow{2}*{22.10}&\multirow{2}*{26.17}&\multirow{2}*{98.68}&\multirow{2}*{98.60} &\multirow{2}*{59.20}&\multirow{2}*{74.28}&\multirow{2}*{56.90}&\multirow{2}*{53.16} &\multirow{2}*{32.83}&\multirow{2}*{18.05}&\multirow{2}*{61.51}&\multirow{2}*{46.61} &\multirow{2}*{\textbf{55.02}} &\multirow{2}*{76.45}    \\
  \specialrule{0em}{-0.5pt}{-2pt}
  {\small \textit{(iteration 2)}}&\multirow{2}*{}&\multirow{2}*{}&\multirow{2}*{}&\multirow{2}*{}&\multirow{2}*{}&\multirow{2}*{}&\multirow{2}*{}&\multirow{2}*{}&\multirow{2}*{}&\multirow{2}*{}&\multirow{2}*{}&\multirow{2}*{}&\multirow{2}*{}\\
  \hline
  \hline
  Chance  &16.50 &16.45 &16.42 &16.65 &16.65 &16.51 &16.07 &16.06 &16.80 &16.74 &16.20 &16.25&16.44 &-\\
  CNN &21.32  &27.89 &15.99 &14.48 &47.81 &50.82 &0.00 &0.00 &19.21 &23.10 &11.70 &2.80 &23.05 &40.32\\
  CNN+LSTM  &22.10  &11.78 &18.55 &16.37 &64.24 &59.57 &13.69 &18.55 &22.70 &29.13 &17.18 &3.61 &24.65 &45.50\\
  CNN+SVM  &19.92  &23.63 &28.46 &38.30 &68.53 &76.07 &15.15 &6.32 &23.28 &16.87 &40.76 &49.24 &36.23 &-\\
  CNN+RF &53.12 &57.98 &20.78 &0.24 &0.00 &0.00 &51.88 &27.31 &15.90 &19.39 &35.56 &44.42 &37.68 &-\\
  \hline
  PRNet  &0.00 &0.00 &47.52 &52.54 &89.63 &58.00 &19.49 &21.52 &19.72 &22.05 &48.69 &62.40 &39.59 &61.45\\
  VGG16 &35.55 &48.93 &99.70 &99.85 &76.95 &13.04 &37.02 &31.88 &26.62 &20.89 &53.05 &59.88 &49.91 &72.18\\
  Resnet50 (192-d) &33.61 &38.19 &78.22 &85.66 &62.27 &76.75 &18.58 &11.21 &35.89 &18.55 &57.82 &60.26 &53.72 &77.16 \\
  
  \hline
  AdjMat-only &34.00 &22.63 &31.46 &22.81 &38.06 &52.42 &27.70 &26.79 &25.42 &25.25 &32.32 &28.69 &32.64 & 46.48\\
  2 branch-iteration 2  &20.43 &8.93 &92.65 &76.03 &47.57 &59.47 &40.34 &45.35 &36.36 &35.77 &55.15 &57.93 & 49.57 & 80.33 \\
  2 branch-iteration 3  & 18.92 & 19.67 & 99.72 & 97.18 & 57.69 & 60.18 & 11.92 & 6.19 & 31.10 & 20.40 & 39.67 & 53.22 & 46.39 & 66.77 \\
 \hline
  Ours-iteration 1  &6.69 &4.66 &49.39 &47.96 &36.56 &39.44 &25.89 &27.82 &35.05 &31.93 &36.71 &42.22 &33.67 &53.97\\
  Ours-iteration 3 &44.83 &0.77 &51.29 &66.41 &47.09 &64.03 &0.00 &0.00 &25.95 &26.20 &47.42 &46.74 &44.52 &72.77 \\
  Ours-iteration 4 &28.01 &5.77 &99.59 &93.15 &42.06 &59.06 &38.46 &14.02 &22.02 &17.54 &43.69 &55.77 &48.35 &72.35\\
Ours w/o. temporal reason.  &13.74 &10.80 &98.64 &98.54 &54.54 &53.17 &55.87 &53.75 &40.83 &25.00 &45.89 &61.55 & 53.73 & 80.33\\
Ours w. implicit learn.  & 30.60 & 9.15 &33.00 &34.56 &43.39 &56.00 &21.50 &26.98 &22.43 &18.63 &58.30 &39.33 & 33.74 &56.54\\
  \hline
  \end{tabular}
  }
  \vspace*{-8pt}
  \caption{\small\textbf{Quantitative results of atomic-level gaze communication prediction}. 
  The best scores are marked in \textbf{bold}. 
  }\label{tab:atomic}
  \vspace*{-10pt}
\end{table*}

Before going deep into our model implementation, we offer a short summary of the whole spatio-temporal reasoning process. As shown in Fig.~\ref{fig:model}, with an input social video (a), for each frame, we build an initial complete graph $\mathcal{G}$ (b) to represent the gaze communication entities (\ie, humans and social scene) by nodes and their relations by edges. During the spatial reasoning step (c), we first update edge representations using Eq.~\ref{eq:1} (note the
changed edge color compared to (b)). Then, in the graph structure update phase, we infer the graph structure through updating the connectivities between each node pairs using Eq.~\ref{eq:2} (note the
changed edge thickness compared to (b)). In the node update phase, we update node embeddings using Eq.~\ref{eq:3} (note the
changed node color compared to (b)). Iterating above processes leads to efficient message propagation in spatial domain. After several spatial message passing iterations, we feed the enhanced node feature into a LSTM based temporal reasoning module, to capture the temporal dynamics (Eq.~\ref{eq:4}) and predict final atomic gaze communication labels (Eq.~\ref{eq:5}). We then use event network to reason about event-level labels based on previous inferred atomic-level label compositions for a long sequence in a larger time scale.

\vspace{-5pt}
\subsection{Detailed Network Architecture}\label{sec:id}
\vspace{-3pt}

\noindent\textbf{Attention Graph Learning}.
In our social graph, the adjacency matrix $\textbf{A}$ stores the attention relations between nodes, \ie, representing the interactions between the entities in the social scene. Since we have annotated all the directed human-human interactions and human-scene relations (\S\ref{sec:DA}), we learn the adjacency matrix $\textbf{A}$ in an explicit manner (under the supervision of ground-truth). Additionally, for the scene node $s$, since it's a `dummy' node, we enforce $a_{v,s}$ as 0, where $v\!\in\!\mathcal{V}$. In this way, other human nodes cannot influence the state of the scene node during message passing.
In our experiments, we will offer more detailed results regarding learning $\textbf{A}$ in an implicit (\textit{w/o.} ground-truth) or explicit manner. 

\vspace{-2pt}

\noindent\textbf{Node/Edge Feature Initialization}. For each node $v\!\in\!\mathcal{V}\backslash s$, the 4096-$d$ features (from the \textit{fc7} layer of a pre-trained ResNet50~\cite{he2016deep}) are extracted from the corresponding bounding box as its initial feature $\mathbf{x}_{v}$. For the scene node $s$, the \textit{fc7} feature of the whole frame is used as its node representation $\mathbf{x}_{s}$. To decrease the amount of parameter, we use fully connected layer to compress all the node features into $6$-$d$ and then encode a $6$-$d$ node position info with it. For an edge $e\!=\!(v,w)\!\in\!\mathcal{V}$, we just concatenate the related two node features as its initial feature $\mathbf{x}_{v,w}$. Thus, we have $V\!=\!12$ and $E\!=\!24$.

\vspace{-2pt}

\noindent\textbf{Graph Network Implementations}. 
The functions $f(\cdot)$ in Eqs.~\ref{eq:1}, \ref{eq:2} and \ref{eq:5} are all implemented by fully connected layers, whose configurations can be determined according to their corresponding definitions. The function in Eq.~\ref{eq:3} is implemented by gated recurrent unit (GRU) network.  

\noindent\textbf{Loss functions}. When explicitly learning the adjacency matrix, we treat it as a binary classification problem and use the \textit{cross entropy} loss. We also employ standard \textit{cross entropy} loss for the multi-class classification of gaze communication labels.

\vspace{-8pt}
\section{Experiments}\label{sec:ex}
\vspace{-3pt}

\begin{table*}[t!]
  \centering
  \resizebox{0.99\textwidth}{!}{
    \setlength\tabcolsep{6pt}
    \renewcommand\arraystretch{1}
  \begin{tabular}{r||c|c|c|c|c|c|c|c|c|c|c|c}
  \hline\thickhline
  \rowcolor{mygray}
     & \multicolumn{12}{c}{Event-level Gaze Communication (Precision  \&   F1-score)}\\
     \cline{2-13}
     \rowcolor{mygray}
   \multirow{-2}{*}{Task~~~~~}&\multicolumn{2}{c}{\textit{Non-Comm.}}&\multicolumn{2}{c}{\textit{Mutual Gaze}}
   &\multicolumn{2}{c}{\textit{Gaze Aversion}}&\multicolumn{2}{c}{\textit{Gaze Following}}&\multicolumn{2}{c}{\textit{Joint Attention}}
   &\multicolumn{2}{c}{\textit{Avg. Acc.}}\\
   \hline
  Metric~~~~ &$\mathcal{P}$ (\%) $\uparrow$ &$\mathcal{F}$ (\%) $\uparrow$  &$\mathcal{P}$ (\%)  $\uparrow$ &$\mathcal{F}$ (\%) $\uparrow$ &$\mathcal{P}$ (\%) $\uparrow$ &$\mathcal{F}$ (\%) $\uparrow$ &$\mathcal{P}$ (\%) $\uparrow$ &$\mathcal{F}$ (\%) $\uparrow$ &$\mathcal{P}$ (\%) $\uparrow$ &$\mathcal{F}$ (\%) $\uparrow$ & top-1 (\%) $\uparrow$ & top-2 (\%) $\uparrow$\\
  \hline
  \hline

Chance & 21.3 & 29.3 & 25.0 & 23.0 & 20.0 & 14.8 & 36.3 & 15.1 & 20.3 & 22.1 & 22.7 & 45.0\\
  \hline
FC-w/o. GT & 43.7 & 44.3 & 16.9 & 23.3 & 6.2 & 10.0 & 8.3 & 9.1 & 60.9 & 40.2 & 35.6 & 69.1 \\
Ours-w/o. GT & 50.7 & 49.3 & 16.7 & 21.0 & 8.2 & 11.3 & 6.2 & 7.7 & 60.9 & 40.0 & \textbf{37.1} & 65.5 \\
  \hline
FC-w. GT & 90.7 & 70.7 & 12.3 & 30.8 & 22.2 & 30.8 & 15.0 & 48.3 & 56.8 & 57.1 & 52.6 & 86.5 \\
Ours-w. GT & 91.4 & 72.7 & 14.5 & 32.3 & 18.5 & 45.5 & 20.0 & 66.7 & 62.2 & 30.8 & \textbf{55.9} & 79.4\\
\hline
\end{tabular}
}
  \vspace*{-8pt}
  \caption{\small\textbf{Quantitative results of event-level gaze communication prediction}. 
The best scores are marked in \textbf{bold}. 
}\label{tab:event}
  \vspace*{-10pt}
\end{table*}

\begin{figure*}
	\centering
   \includegraphics[width=1\linewidth]{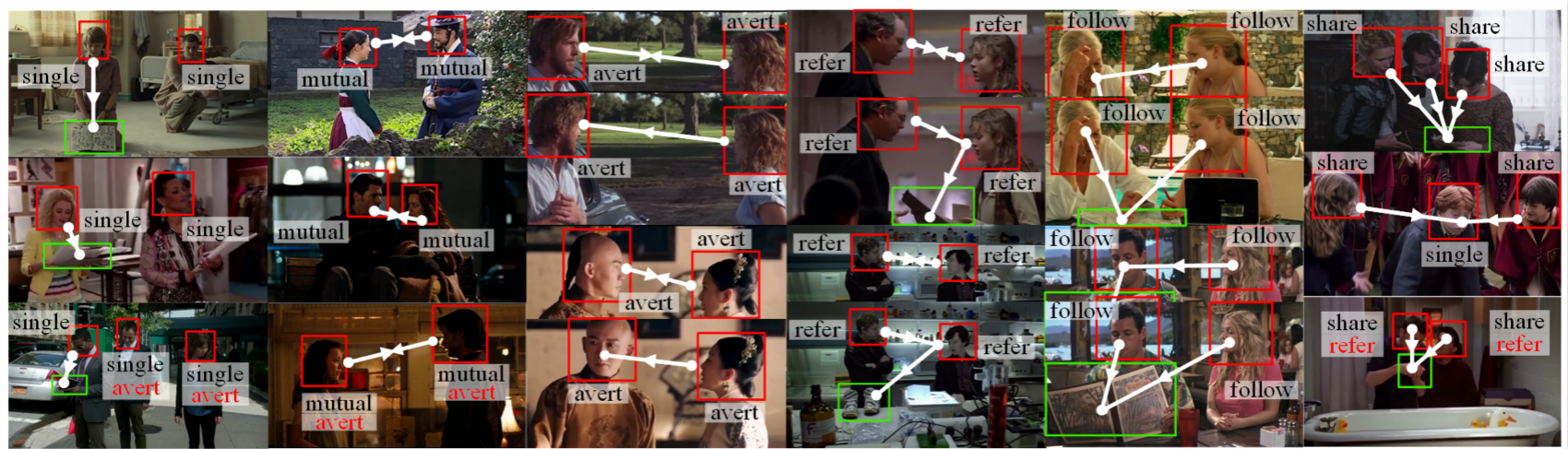}
   \vspace*{-18pt}
   \caption{\small\textbf{Qualitative results of atomic-level gaze communication prediction}. Correctly inferred labels are shown in black while error examples are shown in red.}
\label{fig:qua_res}
\vspace*{-12pt}
\end{figure*}

\subsection{Experimental Setup} \label{subsec:setup}
\vspace{-5pt}
\noindent\textbf{Evaluation Metrics}. Four evaluation metrics, we use precision, F1-score, top-1 Avg. Acc. and top-2 Avg. Acc. in our experiments. Precision $\mathcal{P}$ refers to the ratio of true-positive classifications to all positive classifications. F1-score $\mathcal{F}$ is the harmonic mean of the precision and recall:\! $2\!\times\!\text{precision}\!\times\!\text{recall}/(\text{precision}\!+\!\text{recall})$. Top-1 Avg. Acc. and top-2 Avg. Acc. calculate the average label classification accuracy over all the test set.

\noindent\textbf{Implementation Details}. Our model is implemented by PyTorch. During training phase, the learning rate is set to 1e-1, and decays by 0.1 per epoch. 
For the atomic-gaze interaction temporal reasoning module, we set the sequential length to 5 frames according to our dataset statistics. The training process takes about 10 epochs (5 hours) to roughly converge with an NVIDIA TITAN X GPU. 

\noindent\textbf{Baselines}. To better evaluate the performance of our model, we consider the following baselines:

\noindent{\small\textbullet}~\textit{Chance} is a weak baseline, \ie, randomly assigning an atomic gaze communication label to each human node.

\noindent{\small\textbullet}~\textit{CNN} uses three \textit{Conv2d} layers to extract features for each human node and concatenates the features with position info. for label classification (no spatial communication structure, no temporal relations). 


\noindent{\small\textbullet}~\textit{CNN+LSTM} feeds the CNN-based node feature to an LSTM (only temporal dynamics, no spatial structures).

\noindent{\small\textbullet}~\textit{CNN+SVM} concatenates the CNN-based node features and feeds it into a Support Vector Machine classifier.

\noindent{\small\textbullet}~\textit{CNN+RF} replaces the above SVM classifier with a Random Forest classifier.

\noindent{\small\textbullet}~\textit{FC-w/o. GT \& FC-w. GT} are fully connected layers without or with ground truth atomic gaze labels.

\noindent\textbf{Ablation Study}. To assess the effectiveness of our essential model components, we derive the following variants:

\noindent{\small\textbullet}~\textit{Different node feature}. We try different ways to extract node features. \textit{PRNet} uses 68 3D face keypoints extracted by PRNet~\cite{feng2018prn}. \textit{VGG16} replaces Resnet50 with VGG16~\cite{simonyan2014very}. \textit{Resnet50 (192-d)} compresses the 4096-d features from fc7 layer of Resnet50~\cite{he2016deep} to 192-d.

\noindent{\small\textbullet}~\textit{AdjMat-only} directly feeds the explicitly learned adjacency matrix into some \textit{Conv3d} layers for classification.

\noindent{\small\textbullet}~\textit{2 branch} concatenates a second adjacency matrix branch alongside the GNN branch for classification. We test with different message passing iterations. 

\noindent{\small\textbullet}~\textit{Ours-iteration {1,2,3,4}} test different message passing iterations in the spatial reasoning phase of our full model.

\noindent{\small\textbullet}~\textit{Ours w/o. temporal reason.} replaces LSTM with \textit{Cond3d} layers in the temporal reasoning phase of our full model.

\noindent{\small\textbullet}~\textit{Ours w. implicit learn.} is achieved by unsupervisedly learning adjacent matrix $\mathbf{A}$ (\textit{w/o.} attention ground truths).

\vspace*{-3pt}
\subsection{Results and Analyses}\label{sec:qq}
\vspace*{-3pt}
\noindent\textbf{Overall Quantitative Results}. The quantitative results are shown in Table~\ref{tab:atomic} and \ref{tab:event} respectively for the atomic-level and event-level gaze communication classification experiments. For the atomic-level task, our full model achieves the best top-1 avg. acc. ($55.02 \%$) on the test set and shows good and balanced performance for each atomic type instead of overfitting to certain categories. For the event-level task, our event network improves the top-1 avg. acc. on the test set, achieving $37.1 \%$ with the predicted atomic labels and $55.9 \%$ with the ground truth atomic labels.

\noindent\textbf{In-depth Analyses}.
For atomic-level task, we examined different ways to extract node features and find Restnet50 the best. Also, compressing the Resnet50 feature to a low dimension still performs well and efficiently (full model vs. Resnet50 192-d). The performance of \textit{AdjMat-only} which directly uses the concatenated adjacency matrix can obtain some reasonable results compared to the weak baselines but not good enough, which is probably because that gaze communication dynamic understanding is not simply about geometric attention relations, but also depends on a deep and comprehensive understanding of spatial-temporal scene context. We examine the effect of iterative message passing and find it is able to gradually improve the performance in general. But with iterations increased to a certain extent, the performance drops slightly.

\noindent\textbf{Qualitative Results}.
Fig.~\ref{fig:qua_res} shows some visual results of our full model for atomic-level gaze communication recognition. The predicted communication structures are shown with bounding boxes and arrows. Our method can correctly recognize different atomic-level gaze communication types (shown in black) with effective spatial-temporal graph reasoning. We also present some failure cases (shown in red), which may be due to the ambiguity and subtlety of gaze interactions, and the illegibility of eyes. Also, the shift between gaze phases could be fast and some phases are very short, making it hard to recognize.

\vspace*{-10pt}
\section{Conclusion}\label{sec:conc}
\vspace*{-3pt}
We address a new problem of inferring human gaze communication from both atomic-level and event-level in third-person social videos. We propose a new video dataset \ourdataset~and a spatial-temporal graph reasoning model, and show benchmark results on our dataset. Our model inherits the complementary advantages of graphs and standard feedforward neural networks, which naturally captures gaze patterns and provides better compositionality. We hope our work will serve as important resources to facilitate future studies related to this important topic.  



{\small\noindent\textbf{Acknowledgements} The authors thank Prof. Tao Gao, Tianmin Shu, Siyuan Qi and Keze Wang from UCLA VCLA Lab for helpful comments on this work. This work was supported by ONR MURI project N00014-16-1-2007, ONR Robotics project N00014- 19-1-2153, DARPA XAI grant N66001-17-2-4029, ARO grant W911NF1810296, CCF-Tencent Open Fund and Zhijiang Lab's International Talent Fund for Young Professionals. We gratefully
acknowledge the support of NVIDIA Corporation with the donation of the Titan Xp GPU used for this research.}

{\small
\bibliographystyle{ieee_fullname}
\bibliography{egbib}
}

\end{document}